# Branched Variational Autoencoder Classifiers

Ahmed Salah[*], David Yevick

*Department of Physics and Astronomy, University of Waterloo, ON N2L 3G1, Canada*

*Corresponding author email address: asalah@uwaterloo.ca

***Abstract–*** This paper introduces a modified variational autoencoder (VAEs) that contains an additional neural network branch. The resulting "branched VAE" (BVAE) contributes a classification component based on the class labels to the total loss and therefore imparts categorical information to the latent representation. As a result, the latent space distributions of the input classes are separated and ordered, thereby enhancing the classification accuracy. The degree of improvement is quantified by numerical calculations employing the benchmark MNIST dataset for both unrotated and rotated digits. The proposed technique is then compared to and then incorporated into a VAE with fixed output distributions. This procedure is found to yield improved performance for a wide range of output distributions.

## 1. Introduction

Clustering algorithms, which sort similar data samples into segregated groups, are employed in numerous practical applications involving complex and often noisy data, such as false news detection [1] and document analysis [2]. In such cases, meaningful features for sample assignment are often identified by either of two clustering procedures: similarity-based clustering such as spectral clustering, which involves computing a distance matrix [3], and feature-based clustering represented by K-means and Gaussian mixture models, which instead minimizes the sum of the squared errors between the feature values of the data points and the centroid of the cluster to which they belong.

While at first sight unrelated to clustering algorithms, deep generative models such as variational autoencoders (VAE) employ neural networks to organize input data in a manner that enables the subsequent creation of synthetic samples that exhibit or interpolate the features of the input data [4]. In particular, the high-dimensional input data is encoded through a neural network into a few dimensional latent variable space. The latent variables are then passed through a decoder neural network that generates an output distribution in the original high-dimensional space. Standard VAEs minimize the difference between this output distribution and the input distribution through variational inference. This procedure can compensate for insufficient data and unbalanced labels and has found applications in diverse fields [7] including intrusion detection [5] and target recognition [6].

This paper, however, instead focuses on the ability of the autoencoder latent data representation to effect clustering. Prior work involving clustering that have incorporated autoencoders have included computer vision applications, pattern recognition [8, 9] speech and audio recognition [10, 11] wireless communication [12] and text classification [13]. These examples typically apply clustering algorithms, such as K-means, to the latent variables of the autoencoder [14]. However, the distribution of classes in the latent space may not always be suitable for such procedures.

Numerous generative models and clustering techniques for deep neural networks currently exist. The Variational Fair Autoencoder (VFAE) enhances the separation between latent variables and noise by introducing a regularization term based on the maximum mean discrepancy into the loss function [15]. The Deep Clustering Network (DCN) [14] employs a gradient descent-based formula to address numerical issues with clustering centers. Deep Embedded Clustering (DEC), [16] jointly learns cluster assignments and feature representations through deep neural networks. However, DEC, like K-means, cannot model the data generation process and, therefore, cannot generate new, synthetic, samples.

In contrast, neural network-based clustering models can learn high-quality representations that capture data characteristics and can then generate new data samples, thus combining the strengths of deep clustering and generative models. For example, Tian et al. [17] introduced a comprehensive clustering framework that uses the Alternating Direction of Multiplier Method (ADMM) to update clustering parameters. The Information Maximizing Variational Autoencoder (IM-VAE) instead both increases the mutual information between the latent variables and the samples and minimizes the divergence between the approximate posterior and the true posterior distribution [18]. The Nouveau Variational Autoencoder (NVAE) employs deep separable convolution and batch normalization to enhance the quality of the generated samples [19].

However, the features generated by traditional autoencoders rely on unsupervised techniques that do not incorporate label information and are therefore not optimized for classification. Therefore, when augmenting data for classification purposes, the class labels must be present during feature extraction to ensure that the latent space is discriminative.

To address this limitation, researchers have introduced supervised or semi-supervised autoencoders. For instance, Gao et al. [20] implemented a supervised autoencoder (AE) for face recognition by incorporating a similarity preserving component into the AE's objective function, ensuring that images of the same person are treated as similar. Another approach, which was shown to distinguish rotated digits accurately through their latent space representations, implemented an objective function that evaluates the output linked to each rotated digit by comparing it to a fixed reference digit [21]. Similarly, the Conditional Variational Autoencoder (CVAE) [22] incorporates one-hot encoded labels in order to utilize category information as a control mechanism. Abbasnejad [23] implemented semi-supervised classification by implementing a Dirichlet process that dynamically adjusted the mixing coefficients of a combination of VAEs according to the properties of the input data. The Orthogonal AutoEncoder (OAE) further ensures the orthogonality of the resulting embedding while the Clustering framework based on Orthogonal AutoEncoder (COAE) additionally enables both the extraction of latent embeddings and the generation of clusters [24]. A recent method introduced by Song et al. [25] combines both the reconstruction error and the error derived from comparing K-means clustering with the encoded image into a single objective function. In this manner, the dissimilarity between the original latent space learned by AE and the feature space derived from it using traditional machine learning (ML) techniques is minimized. Similarly, The Fisher Variational Autoencoder (FVAE) integrates the Fisher criterion into the VAE by incorporating the Fisher regularization term into the loss function with the aid of the class labels [26]. By maximizing the distance between different classes and minimizing the distance within the same class the latent variables can be more accurately classified. A similar approach involving the addition of a supervised technique to the VAE to improve classification accuracy will be introduced below.

In this work, we introduce a novel regularized Variational Autoencoder known as the Branched Variational Autoencoder (BVAE) in order to improve the identification of different classes from the associated clusters in the VAE latent space. This is achieved by processing the inputs within each cluster with traditional machine learning methods that are applied to a secondary classifier branch. The objective function then combines two elements, the reconstruction error of the variational autoencoder and the classifier branch loss term. Both the data representation and the classifier loss are iteratively updated. The resulting procedure is applicable to a broad range of datasets as will be demonstrated in the context of the MNIST dataset by employing the procedure to enhance the performance of the K-Means clustering algorithm as well as to identify randomly rotated digits.

## 2. Variational Autoencoders

Variational Autoencoders are generative models that represent the actual distribution of data samples by a low-dimensional approximate "latent variable" distribution However, VAEs assume that the latent variable distribution is continuous and follows a normal distribution. This assumption does not necessarily reflect the true distribution of complex data, leading to a mismatch between the assumed and actual distributions. In addition, different classes may not be effectively separated in the latent space under this assumption, especially when two distinct classes have very close mean and variance values [27-29]. In the VAE, the latent space variables are mapped to an image in the space of the input variables as indicated in Fig. 1 which depicts the VAE architecture. In this model, $z$ represents the latent variable, while $\mu$ and $\sigma$ denote the mean and standard deviation of $z$, respectively. After the latent space is determined, mapping any point in the latent space back to the image space generates a novel image. The conditional distributions, $q_\phi(z|x)$ and $p_\theta(x|z)$ that are learned by the encoder and decoder, respectively are termed recognition and generation models, while $\varphi$ and $\theta$ represent the corresponding model parameters. The VAE typically employs Gaussian distributions with diagonal covariances for both the encoder $q_\varphi(z|x)$ and decoder $p_\theta(x|z)$. The estimated posterior distribution $q_\varphi(z|x)$ is utilized to approximate the unknown prior distribution $p(z)$ that represents the distribution of latent variables $z$ in the absence of any specific input data $x$.

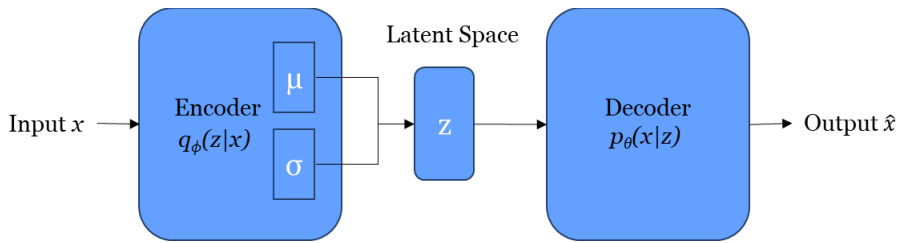

Fig. 1 Variational autoencoder structure

One main issue is that the marginal likelihood given by

$$p_\theta(x) = \int p_\theta(x, z)\, dz \tag{1}$$

is intractable as the integral does not have an analytic solution [30]. This intractability is related to the intractability of $p_\theta(z|x)$, where

$$p_\theta(z|x) = \frac{p_\theta(x, z)}{p_\theta(x)} = \frac{p_\theta(x|z)\, p_\theta(z)}{p_\theta(x)} \tag{2}$$

Furthermore $p(z)$ cannot be directly estimated [29]. To tackle the intractability problem, the posterior distribution $q_\varphi(z|x)$ is introduced where $\varphi$ refers to the parameters of this inference model that is optimized so that $q_\varphi(z|x) \approx p_\theta(z|x)$.

The optimization objective of the VAE is the evidence lower bound (ELBO) [30], and is realized through maximum likelihood estimation, where, for any $q_\varphi(z|x)$, the log likelihood function of the VAE is

$$\log p_\theta(x) = KL\left(q_\varphi(z|x) \big| p_\theta(z|x)\right) + L_{\theta,\varphi}(x) \tag{3}$$

The first term is the Kullback-Leibler (KL) divergence that quantifies the difference between the distributions $q_\varphi(z|x)$ and $p_\theta(z|x)$,

$$KL\left(q_\varphi(z|x) \big| p_\theta(z|x)\right) = E_{q_\varphi(z|x)}\left[\log\left[\frac{q_\varphi(z|x)}{p_\theta(z|x)}\right]\right] \tag{4}$$

$L_{\theta,\varphi}(x)$, which is the evidence lower bound (ELBO) of the likelihood function can be written as

$$L_{\theta,\varphi}(x) = E_{q_\varphi(z|x)}\left[\log\left(\frac{p_\theta(x,z)}{q_\varphi(z|x)}\right)\right] \tag{5}$$

$$= E_{q_\varphi(z|x)}[\log p_\theta(x|z) + \log p_\theta(z) - \log q_\varphi(z|x)]$$

$$= E_{q_\varphi(z|x)}[\log p_\theta(x|z)] - KL(q_\varphi(z|x) | p_\theta(z))$$

Since the KL divergence is non-negative, the ELBO is a lower bound on the log likelihood of the data as from Eq. (3), $L_{\theta,\varphi}(x) = \log p_\theta(x) - KL\left(q_\varphi(z|x) \big| p_\theta(z|x)\right) \leq \log p_\theta(x)$, hence maximizing $L_{\theta,\varphi}(x)$ maximizes $p_\theta(x)$, and further minimizes the difference between the

approximate $q_\varphi(z|x)$ and the true posterior $p_\theta(z|x)$, therefore the VAE asymptotically minimizes the loss function

$$\text{Loss}_{\text{VAE}} = -E_{q_\varphi(z|x)}[\log p_\theta(x|z)] + KL(q_\varphi(z|x)|p_\theta(z)) \qquad (6)$$

composed of the sum of the negative of the reconstruction error and the KL divergence. If $p_\theta(z)$ is assumed to be Gaussian with $N(0, I)$, and setting $q_\varphi(z|x) = \prod_i N(z_i; \mu_i, \sigma_i^2)$ the KL divergence can be evaluated as [21],

$$KL\left(q_\varphi(z|x)\big|p_\theta(z)\right) = -\frac{1}{2}\sum_{i=1}^{k}(1 + \log\sigma_i^2(x) - \sigma_i^2(x) - \mu_i^2(x)) \qquad (7)$$

Since the sampling of the latent variable $z$ is non-differentiable, however, backpropagation cannot be employed in the gradient descent algorithm which would greatly increase the difficulty of optimizing the network parameters. This problem is circumvented with the reparameterization trick which samples $\epsilon$ in

$$z = \mu + \sigma * \epsilon \qquad (8)$$

from a normal distribution with mean 0 and a diagonal covariance matrix with elements $\sigma$. This effectively transforms the sampling of $z$ into a linear operation enabling backpropagation. As $\epsilon$ is a random variable, any point in proximity to a latent position at which inputs are encoded yields a reconstructed image that resembles the averaged input data mapped to or near the position.

## 3. Experimental results
### 3.1 Structure and Objective Function of the BVAE

The BVAE architecture shown in Fig. 2 consists of two main components: the VAE of Fig. 1 and a classifier branch such as a neural network, k-nearest neighbors, or random forest that samples the latent space of the VAE. The VAE learns latent features, while the classifier branch promotes cluster formation. The latent variables $z$ together with the associated labels are input into classifier in order to compensate for the absence of label information in the standard VAE. The training phase of the BVAE then incorporates the classifier branch loss, $L_C$, defined as the categorical cross entropy between the labels predicted from the classifier branch and the true labels, to promote the clustering of related features.

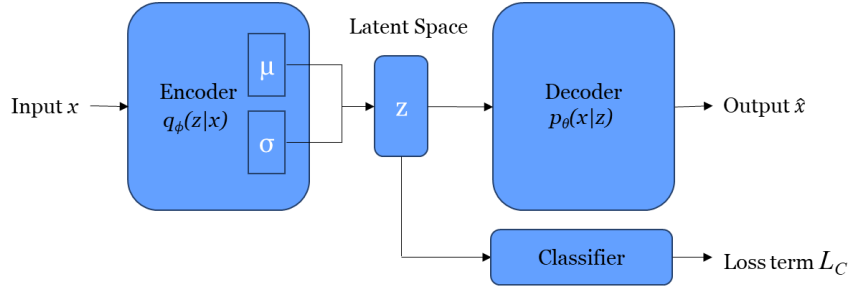

Fig. 2 VAE with a classifier branch

Including $L_C$, the objective function of the BVAE is

$$L_{BVAE} = \alpha L_{con} + L_{KL} + \lambda L_C \tag{9}$$

where $\lambda$ is a regularization parameter that facilitates balancing of the input dataset and $L_{KL}$ and $L_C$ denote the KL divergence and classifier branch loss respectively. The reconstruction loss, $L_{con}$, quantifies the error incurred when the input data is regenerated form the latent space distribution by the VAE, while the regularization term, $\lambda L_C$ incorporates the influence of the prior distribution, which is typically Gaussian on the latent space distribution. A further regularization parameter $\alpha$ is introduced in Eq. (9) to control the relative amplitude of the reconstruction loss. Accordingly, minimizing the loss function in the BVAE, not only reduces the reconstruction error but also the KL divergence as well as the classification error among the latent variables. Note that Eq. (9) yields the standard VAE objective function when $\lambda = 0$.

## 3.2 Training and Optimization

The BVAE implements encoding, sampling, and decoding in the same manner as the standard VAE. The encoder employs the recognition model $q_\varphi(z|x)$ to compute the mean $\mu$ and standard deviation $\sigma$ of the latent variables from the input data $x$. In the sampling step, $\epsilon$ is

generated from a standard normal distribution $N(0, I)$ implemented through the variables $\mu$ and $\sigma$, and the reparameterization trick employed. The decoder finally employs the generative model $p_\theta(x|z)$ to reconstruct the output pattern $\hat{x}$. After training, the generative model $p_\theta(x|z)$ can synthesize new samples from appropriate choices of the latent variables.

The optimization process in VAEs maximizes the likelihood of the observed data, which is termed the evidence lower bound (ELBO) by optimizing the model parameters $\varphi$ and $\theta$ of the loss function through stochastic gradient descent (SGD) and back propagation. The reconstruction loss, KL divergence, and, in the case of the BVAE, the classifier loss term introduces further parameters associated with the weighting of the individual loss terms, all of which must also be minimized. Therefore, the range of acceptable metaparameters in e.g. the BVAE is more restricted than in standard VAE calculations.

## 4. Results and Discussion

### 4.1 Implementation

The proposed approach is implemented by modifying and applying the readily manipulated TensorFlow based code of Chapter 12 of [31] to the standard benchmark MNIST dataset consisting of 70,000 images of handwritten digits discretized as 28×28 arrays of 8-bit pixels. The encoder consists of an input layer, two 2-D convolutional layers of sizes 32 and 64 respectively with 3×3 filter functions, strides = 2, and padding = "same". The resulting 7×7 filters are then flattened and fed to a 16-element dense layer followed by the dense two-dimensional latent space layer. The decoder network consists of a 3136 element dense layer that is equivalent to the product $7 \cdot 7 \cdot 64$ to later create a feature map that is 7 units wide and 7 units high and depth of 64 (number of channels), so it is then reshaped into a $7 \times 7 \times 64$ tensor and fed into two 2-D transpose convolutional layers with 64 and 32 filters respectively each of size 3×3 and finally a 2-D 3×3 filter layer that compresses the information present in the filter outputs from the second transpose layer into the 28×28 matrix reconstructed image. All layers employ **RELU** activation functions except for the standard **sigmoid** final layer. The VAE is trained with the Adam optimizer until convergence is attained at 30 epochs for a batch size of 512.

## 4.2 Clustering Metrics

The classification accuracy is determined by inserting the VAE latent variables of each test data record and label into a NN, while the degree of clustering is quantified with either the Normalized Mutual Information (NMI), Accuracy (ACC) or the adjusted Rand Index (ARI) procedures. In particular,

- The NMI quantifies the similarity between pairs of clusters. In terms of information theory,

$$NMI = \frac{MI(c,l)}{\max(H(c), H(l))}$$

  where $MI(c,l)$ denotes the mutual information between the predicted clusters ($c$) and the ground truth labels ($l$), while $H$ denotes the entropy.

- The ACC determines the mean accuracy based on the alignment between the ground truth labels and the predicted assignments according to

$$ACC = \max_m \frac{\sum_{i=1}^{N} 1\{l_i = m(c_i)\}}{N}$$

  in which $l_i$ represents the true label, $c_i$ is the clustering assignment and the index $m$ ranges over all possible one-to-one mappings between $c_i$ and $l_i$.

- The ARI measures the correspondence between the true labels and the predicted clusters by counting pairs that are assigned to either the same or to different clusters as follows

$$ARI = \frac{RI - E(RI)}{\max(RI) - E(RI)}$$

  Here $RI = (a+b)/\,_nC_2$ is a random index which yields an estimate of the degree of resemblance of two data clusters, where a and b refer to the number of pairs assigned to identical and different clusters respectively and $_nC_2$ is a combinatorial coefficient.

The clusters are determined by applying the $k$-means procedure to the latent variables of the VAE. However, the $k$-means procedure assumes spherical clusters while the actual digit clusters can be elongated, reducing the $k$-means accuracy compared to metrics based on the NN performance.

### 4.3 BVAE with a NN Classifier Branch

As detailed in the previous section, the BVAE integrates the VAE with a NN classifier branch that employs the latent variables and labels as input data. To examine the encoding and clustering of latent variables of different categories, a NN branch consisting of 3 dense layers with 512, 256 and 128 neurons and RELU activation functions followed by a 10-element dense layer with Softmax activation function, will be employed together with a **categorical_crossentropy** loss function. Surprisingly, although NN layers with fewer neurons give less classification accuracy, a single 'linear' layer of 2 neurons gives the same accuracy and performance as that obtained with 3 huge nonlinear layers. Additionally, the weighting factors for the loss terms in the BVAE objective function must be properly chosen. For example, the classifier loss weight, $\lambda$, must be sufficiently large to ensure that the VAE behavior is affected by the classifier branch. The degree of improvement in the clustering is determined by comparing the BVAE with the standard VAE (i.e. the BAE result for $\lambda = 0$). The BVAE will also be compared and subsequently combined with the fixed output VAE method which modifies the VAE objective function such that the cross-entropy relies on predefined target output distributions (10 representative digits chosen from the MNIST dataset in the calculations below), while the full MNIST dataset is again employed as the input data [21].

The efficiency of the BVAE is evident from the clusters in Fig. 3 which displays the 2-D latent variable spaces generated by the MNIST dataset. The digits are distinguished by the greyscale intensity shown on the color bar, such that for example occurrences of 0 and 9 are displayed as black and white dots, respectively. Fig. 3(a) refers to the results of the standard VAE ($\lambda = 0$), while Fig.3(b) displays enhanced latent variable clusters obtained by the VAE with fixed output distributions. However, improved clustering can also be realized by modifying the BAE weighting factors. Employing the BVAE with $\lambda = 100$ yields the clusters of Fig.3(c) while the BVAE with $\alpha = 0.01$ generates distinct clusters with modified shapes as shown in Fig.3(d).

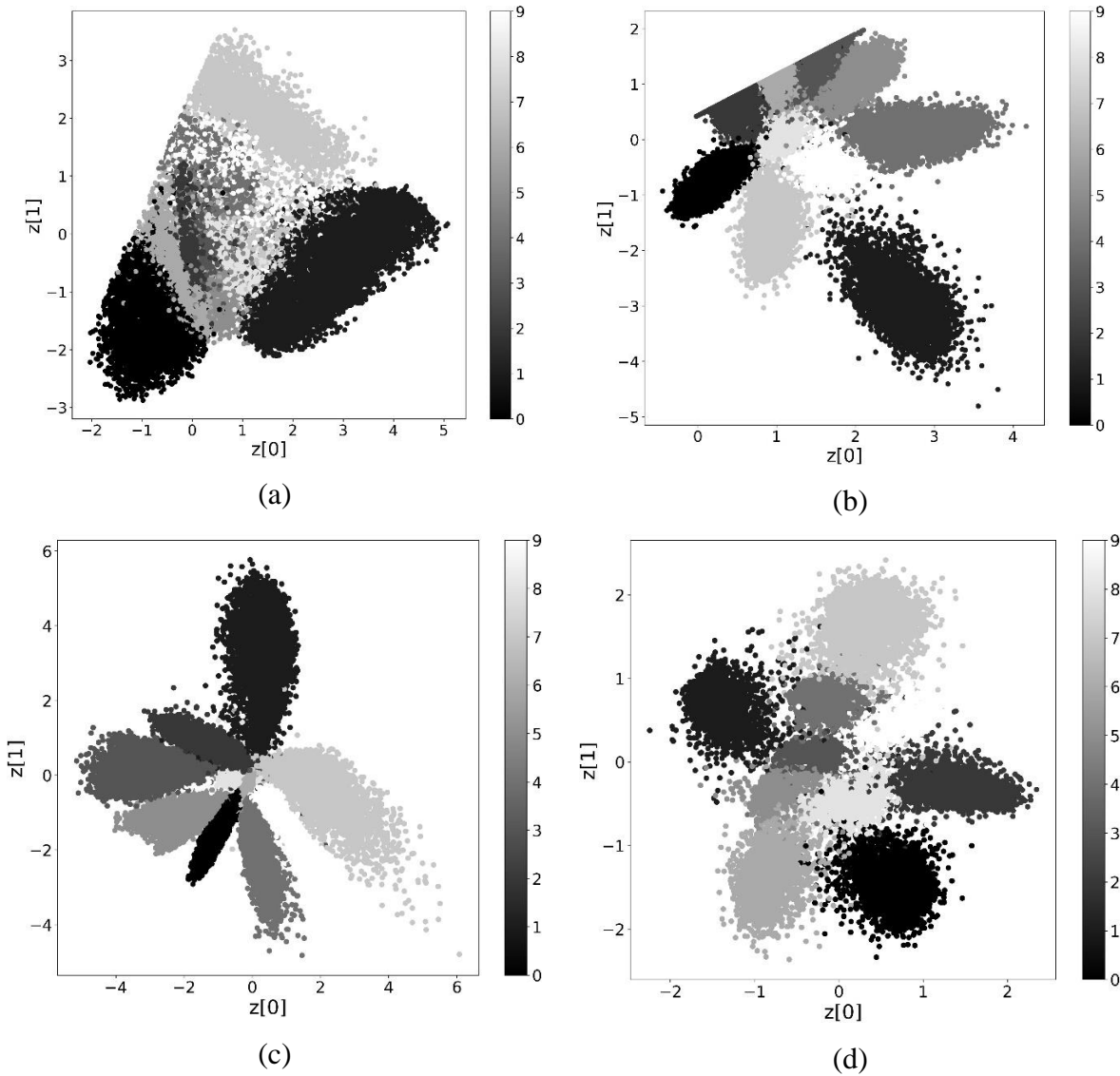

Fig. 3. The 2-D latent variable space distributions generated by the MNIST data set using (a) the standard VAE, (b) the VAE with fixed output distributions (c)-(d) the BVAE with $\lambda = 100$ and $\alpha = 0.01$ respectively.

The reconstructed images from the BVAE are further analyzed in Fig. 4, which displays a 30×30 grid of the patterns produced by the decoder when applied to equidistant latent points from -3 to +3 along both coordinate directions. Fig. 4(a) presents the output patterns for the standard VAE while Fig. 4(b) displays the corresponding result for the BVAE with $\lambda = 100$. The increased isolation of the digit regions in the latent space in the latter case is again evident from the figures.

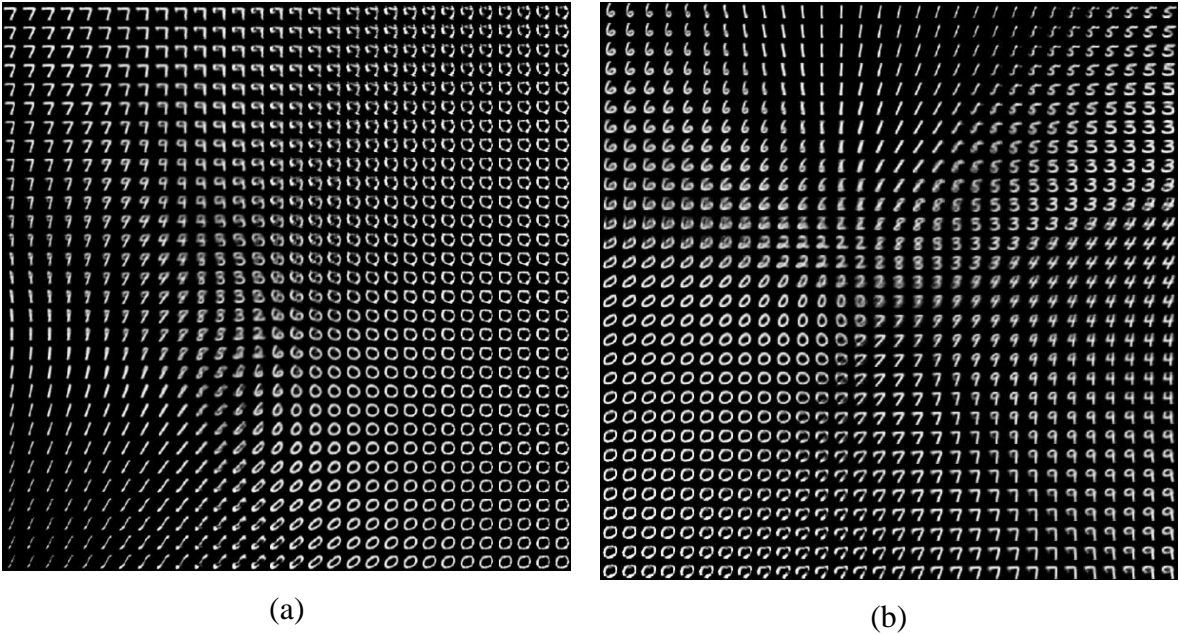

(a) (b)

Fig. 4. The output patterns corresponding to the realizations within a two-dimensional histogram in latent space for (a) the standard VAE, (b) the BVAE with $\lambda = 100$.

To compare the BVAE performance to those of the VAE and the VAE with fixed output, the metrics of the previous section are evaluated, and the results collected in Table 1, where the numbers are averages over independent calculations. Evidently, for $\lambda = 1$, the BVAE does not yield any enhancement compared to the standard VAE. When $\lambda = 10$, however the NN branch slightly influences the VAE behavior while for $\lambda = 100$ the BVAE exhibits an increase in classification accuracy from ≈67% to 98%. While the NMI generates nearly the same degree of enhancement, the enhancement of ACC and ARI is marginally lower. The classification accuracy is almost equivalent to that of the VAE with fixed output, although the other metric values are superior in the fixed output calculation.

The BVAE with $\alpha = 1$ and $\lambda = 100$, exhibits nearly identical classification accuracy but higher values of the remaining metrics in Table 1 when compared to the parameter values $\alpha = 0.01$ and $\lambda = 1$. The larger values of the NMI, ACC and ARI metrics for $\alpha = 1$ and $\lambda = 100$ result largely from the circularity of the clusters of Fig. 3(d) relative to those of Fig. 3(c) since these metrics are based on the *k*-means procedure. Hence, the optimal choice of regularization parameter is somewhat problem-dependent.

Combining both the BVAE with $\lambda = 100$ and the fixed output procedure yields both a superior classification accuracy of 99% and large values for the remaining performance metrics, attesting again to the superior performance that can be achieved with fixed output distributions.

*Table 1 Comparison of clustering performance and classification accuracy for different frameworks. α is the standard VAE loss parameter, while λ parametrizes the influence of the NN classifier branch.*

| Parameter | NMI | ACC | ARI | Classification Accuracy |
| --- | --- | --- | --- | --- |
| VAE | 0.467 | 0.479 | 0.352 | 0.672 |
| VAE + Fixed Output | 0.875 | 0.914 | 0.843 | 0.977 |
| BVAE ($\lambda = 10$) | 0.615 | 0.622 | 0.477 | 0.854 |
| BVAE ($\lambda = 100$) | 0.757 | 0.717 | 0.608 | 0.98 |
| BVAE ($\alpha = 0.01$) | 0.907 | 0.957 | 0.908 | 0.968 |
| BVAE + Fixed Output | 0.854 | 0.867 | 0.802 | 0.99 |

The behavior of the various methods can be quantified by confusion matrices such as those of Fig. 5 which pertain to the NN branch of the BAE. The diagonal elements contain the number of correctly classified instances for each digit, while the off-diagonal elements indicate the number of misclassifications from the digit in the row number to that of the column number. Fig 5(a) and (b) were generated with the standard VAE and the BVAE with $\lambda = 100$, respectively. Evidently, the standard VAE confuses 4 and 9 while predicting the 5 as 3 and 9 as 7 about 500 times. The BVAE with $\lambda = 100$, however, tackles these issues, and the largest number of misidentifications for any of the digits is ≈30 instances.

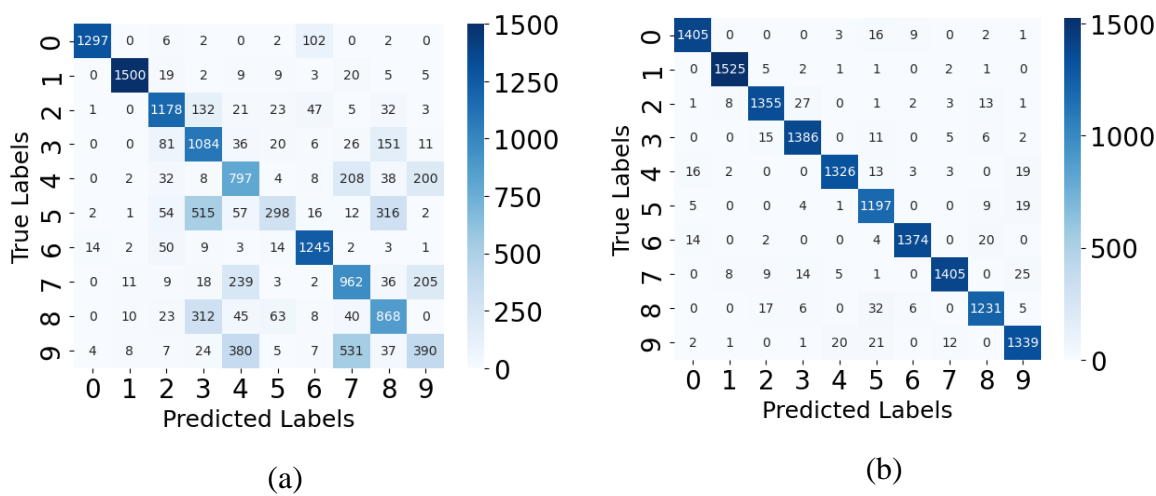

Fig. 5. The confusion matrix associated with the NN accuracy classifier applied to the latent variables for the (a) Standard VAE, (b) BVAE with $\lambda = 100$.

An advantage of the VAE with fixed output framework is that it can be employed with effectively arbitrary output distributions. For example, one possible set of outputs is displayed in the ten pictures of Fig. 6, each of which contains one of 10 Gaussian functions centered at a different position compared to the other images. Each category of input digits (e.g. 0 through 9) is then mapped to the corresponding output position. An additional two sets of output distributions were also generated, one which replaced the Gaussian functions of Fig. 6 with square shapes and the second with a two-dimensional Haar wavelet defined on a square interval. To compare the VAE performance for the three sets to MNIST target outputs, the reconstruction loss is calculated with the **mean_squared_error** rather than the **binary_crossentropy** routine and the sigmoid activation function in the last decoder layer is replaced with a RELU activation. As evident from Fig. 7, which displays the latent space distributions for the (a) MNIST (b) Gaussian (c) square and (d) wavelet outputs, all four output distributions yield highly clustered latent space distributions compared to the standard VAE procedure. Significantly, however, the latent space distributions for cases (b)-(d) are more elongated and isolated than those for the MNIST digits in Fig. 7(a).

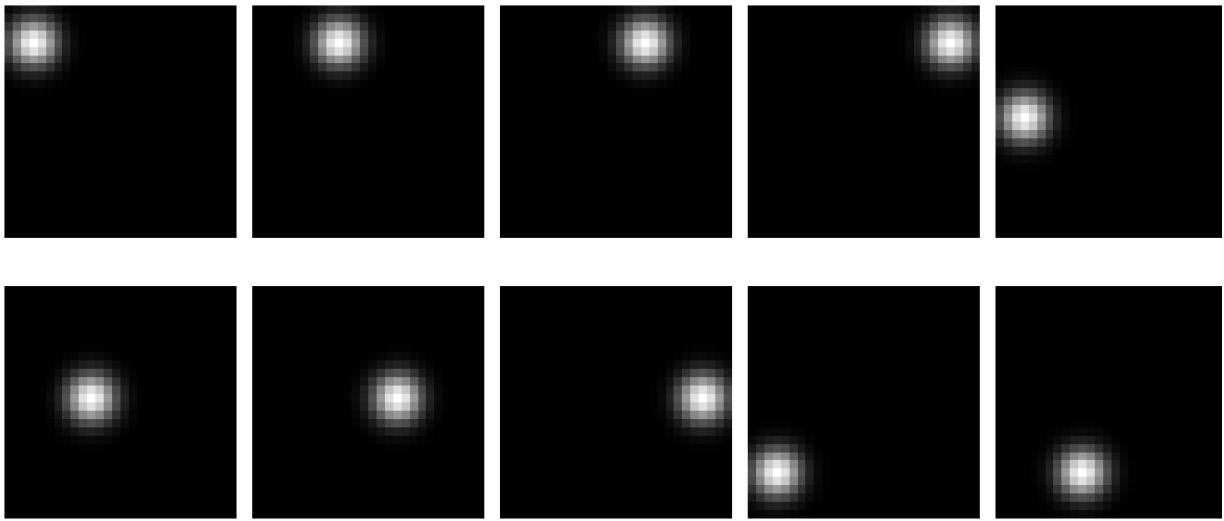

Fig. 6 Synthesized Gaussian fixed output distributions.

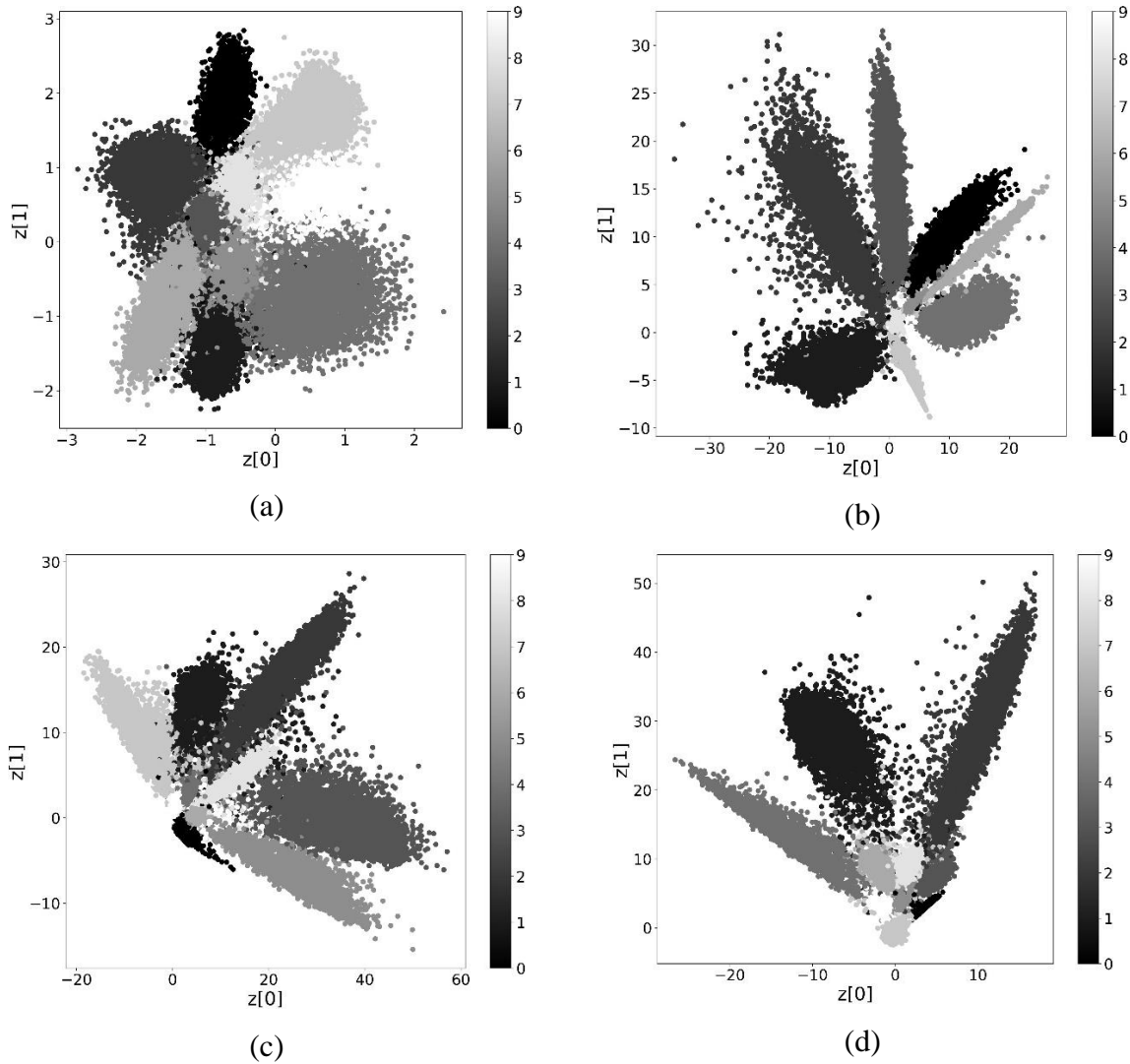

Fig. 7. The 2-D latent variable space distributions for the fixed (a) MNIST (b) Gaussian (c) square and (d) wavelet outputs.

To further assess the VAE with different sets of target output, a sample point from the latent space is passed to the decoder to reconstruct the image of Fig. 8. The result corresponds to a mixture of two Gaussians that encode the digits 9 and 4. This implies that the latent space distributions of these digits share a common boundary, in agreement with the result for the fixed MNIST digit outputs of Fig. 3(b). Further, from Table 2, while all sets of fixed outputs yield nearly the same high degree of classification accuracy, the clustering metrics are largest for the MNIST digit target outputs, presumably because they yield the most circular latent space digit distributions in Fig.7.

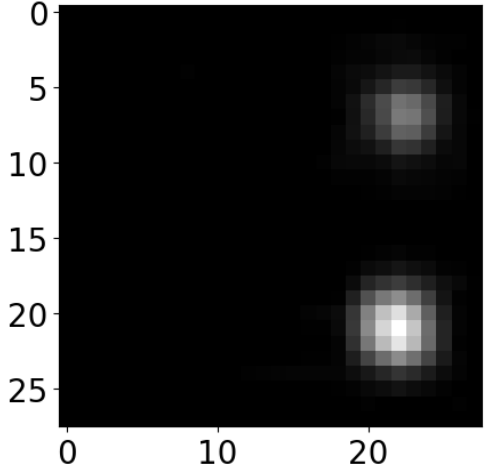

Fig. 8. The reconstructed image generated by a sample point from the 2D latent space for a VAE with Gaussians as fixed target distributions.

*Table 2 Comparison of clustering performance and classification accuracy for the VAE with different sets of fixed outputs.*

| Parameter | NMI | ACC | ARI | Classification Accuracy |
|---|---|---|---|---|
| MNIST | 0.874 | 0.911 | 0.843 | 0.98 |
| Gaussians | 0.82 | 0.807 | 0.732 | 0.972 |
| Squares | 0.793 | 0.754 | 0.674 | 0.977 |
| Wavelets | 0.817 | 0.807 | 0.732 | 0.974 |

**4.4 BVAE with a NN Classifier for Rotated Digits**

To demonstrate the applicability of the proposed architectures to a wide range of classification problems, the randomly rotated MNIST character set will now be employed as input data. Accordingly, Fig. 9 compares the latent variable distributions for rotated digits with both (a) the standard VAE and (b) the BVAE with $\lambda = 100$. Consistent with the results reported in [21], the digit distributions in latent space for randomly rotated digits with different values overlap considerably for the VAE while the digits form identifiable clusters when the BVAE is employed. Indeed, as seen from Table 3, the BVAE with $\lambda = 100$ both significantly increases the clustering metrics and improves the classification accuracy of the VAE from 30% to about 83%, which is comparable with the enhancement obtained with the VAE with the MNIST fixed digit output distributions. At the same time, an NN accuracy of about 87% is achieved when combining the fixed output with the BVAE. While the standard VAE cannot distinguish between most rotated digits, the major source of confusion in the BVAE is limited to 6 and 9 (since these digits yield a nearly identical signature when randomly rotated), and 3 and 8 as other digits are accurately classified, as evident from the confusion matrices in Fig. 10.

Rotated digits are more compartmentalized in latent spaces of higher dimensions which introduce additional degrees of freedom into the network. However, the classification accuracy of the BVAE with a 2-dimentional latent space is comparable to or even exceeds that of the VAE with a 10-dimentional latent space as evident from table 4. Perhaps unexpectedly, while the accuracy of the VAE with fixed output distributions is comparable to that of BVAE for two-dimensional latent spaces, the fixed output VAE is more accurate for higher dimensional latent spaces.

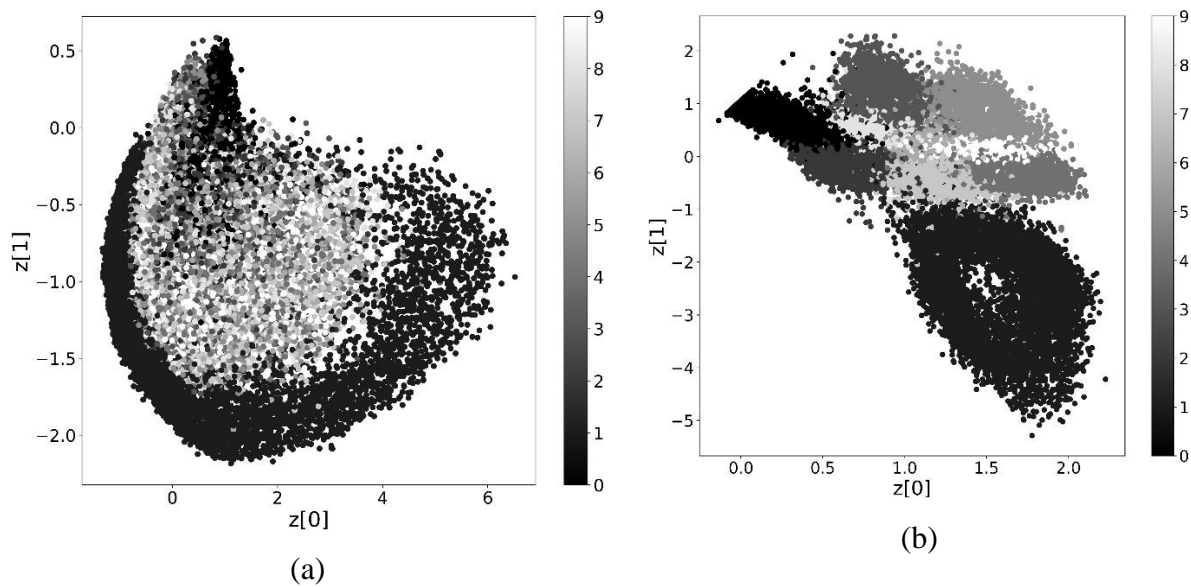

Fig. 9. The two-dimensional latent variable space distribution for randomly rotated MNIST digits in case of the (a) standard VAE (b) BVAE with $\lambda = 100$.

Table 3 Same as table 1 but for rotated MNIST digits.

| Parameter | NMI | ACC | ARI | Classification Accuracy |
|---|---|---|---|---|
| VAE | 0.097 | 0.234 | 0.06 | 0.307 |
| VAE + Fixed Output | 0.646 | 0.718 | 0.551 | 0.837 |
| BVAE ($\lambda = 10$) | 0.223 | 0.304 | 0.14 | 0.435 |
| BVAE ($\lambda = 100$) | 0.571 | 0.589 | 0.434 | 0.833 |
| BVAE ($\alpha = 0.01$) | 0.544 | 0.554 | 0.419 | 0.701 |
| BVAE + Fixed Output | 0.66 | 0.693 | 0.545 | 0.868 |

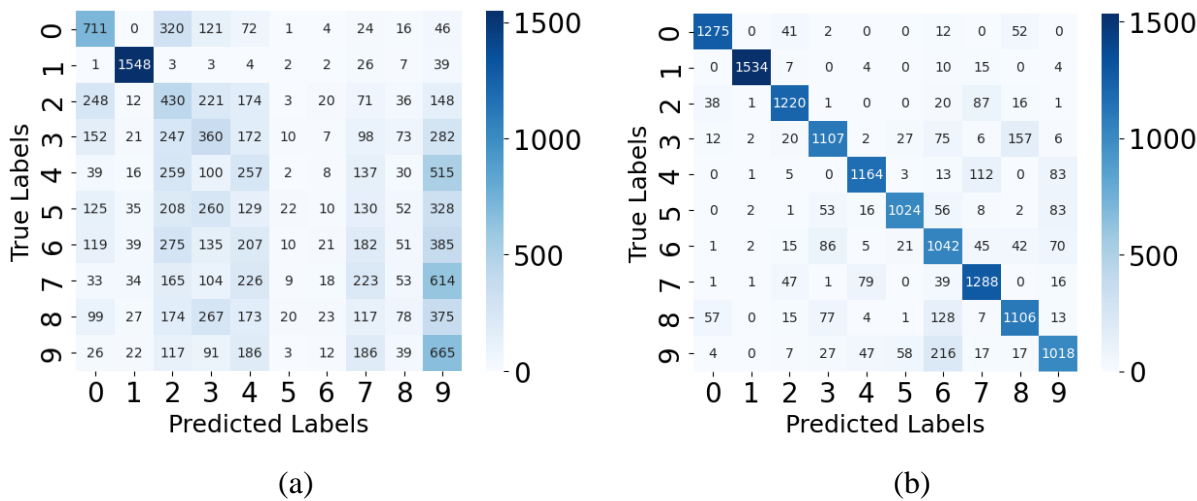

(a)      (b)

Fig. 10. Same as Fig. 5 but for randomly rotated MNIST digits.

Table 4 Classification accuracy for different frameworks with latent spaces of different dimensions and randomly rotated digits.

| Dimension | VAE | VAE + Fixed Output | BVAE |
|---|---|---|---|
| 2 | 0.31 | 0.837 | 0.833 |
| 3 | 0.423 | 0.914 | 0.904 |
| 5 | 0.624 | 0.932 | 0.91 |
| 10 | 0.791 | 0.956 | 0.938 |

## 4.5 BVAE with Classifiers

The NN branch in the BVAE framework can be replaced by any classifier, which introduces additional hyperparmeters. In the case of a *k*-nearest neighbor (*knn*) branch, the hyperparameter is the number, $n$, of nearest neighbors while for the Random Forest (*RF*) $n$ is associated with the number of estimators. These classifiers are here implemented with the **KNeighborsClassifier** and **RandomForestClassifier** routines in the scikit-learn python library. In contrast to the NN branch, a classifier loss factor of $\lambda = 10$ rather than $\lambda = 100$ is found to yield improved performance. Values of $n$ of from 5 to 50 increase the classification accuracy for both the knn and RF methods from 66% to 70% − 73% with minimal further improvement for $n > 50$. However, the enhancement can be increased by altering the weights of the input digits. For example, for a BVAE with $\lambda = 10$ and a *knn* branch with $n = 40$, multiplying the input digit distributions for 0,1 and 2 by a factor of 10 while dividing the digits 3,6,7 and 9 by the same factor increases the accuracy from 73% to 83%. As well, the latent space distribution is considerably affected by the selective nature of the weights as evident from Fig. 11, as the clusters for the digits with smaller weight values are far more distinct. Other weightings can be identified that yield similar performance, for example multiplying the inputs for 0,1,2 and 6 by 2 and dividing those for 3,5,7 and 9 by 2, or alternatively multiplying digits 0 and 6 by 2 and dividing 4,5 and 8 by 2. All such weight combinations, however were found to yield accuracies between 73% and 83%. Similar enhancements can, of course, be realized by weighing the inputs of the standard VAE.

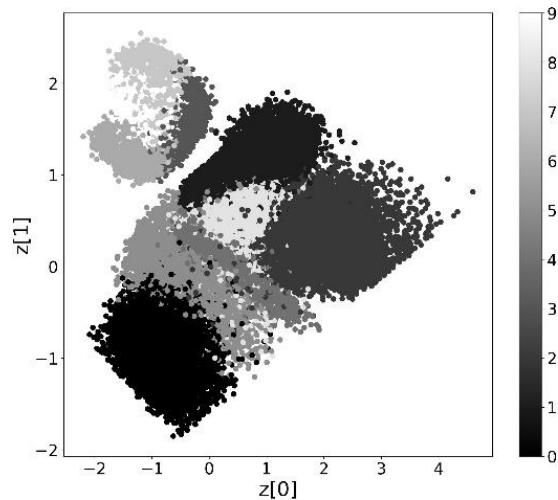

Fig. 11. The 2D latent variable space distribution for a BVAE with $\lambda = 10$ and a *knn* branch for appropriately weighted inputs.

## 5. Conclusions

This paper has advanced a novel regularized variational autoencoder termed the BVAE that significantly enhances the accuracy and the latent variable clustering of the VAE. Further, by incorporating a classifier branch into the VAE, the BVAE transforms unsupervised into supervised learning. This requires redefining the VAE loss function to include the classifier branch loss and therefore introduces an additional hyperparameter corresponding to the magnitude of the additional loss, which when chosen judiciously maximizes the discrimination among classes.

The BVAE with a NN classifier branch and a two-dimensional latent space applied to the standard MNIST dataset yields a classification accuracy of 97% while the corresponding value for the standard VAE is 67% at the same time that other clustering metrics are similarly increased. While the VAE with fixed output distributions offers comparable improvements, combining the BVAE with fixed output distributions yields additional accuracy enhancements. These results are even more pronounced for randomly rotated MNIST input digits where the classification error of the BVAE with a NN branch decreases by a factor of 2.5 compared to that of the of the VAE since the number of occurrences of misclassification of similar digits is greatly reduced. This performance is in fact superior to that of the traditional VAE with a 10-dimentional latent space. Similar results were obtained for the BVAE with a *knn* or *RF* classifier branch while weighting the input data to increase the contribution of the most frequently misinterpreted digits to the loss function yields further accuracy improvements. Consequently, these architectures appear to be promising candidates for numerous practical classification tasks.